\documentclass[preprint,review,12pt]{elsarticle}

\usepackage{amssymb}
\usepackage{amsthm}
\usepackage[figuresright]{rotating}
\usepackage{amsmath}
\usepackage{mathrsfs}
\usepackage{newtxmath}
\usepackage{amsfonts}
\usepackage{mathtools}
\usepackage{algorithm}
\usepackage{algorithmic}
\usepackage{soul}
\usepackage{url}
\usepackage[utf8]{inputenc}
\usepackage{caption}
\usepackage{subcaption}
\usepackage{graphicx}
\usepackage{booktabs}
\usepackage{xcolor}
\usepackage{multirow}
\usepackage{float}
\usepackage{comment}
\newcommand{\etal}{\textit{et al.}}
\newcommand{\ie}{\textit{i.e. }}
\newcommand{\eg}{\textit{e.g. }}
\DeclareMathOperator{\softmax}{softmax}
\usepackage{setspace} 

\journal{Pattern Recognition}
\begin{document}
\begin{frontmatter}

\title{VLCDoC: Vision-Language Contrastive Pre-Training Model for Cross-Modal Document Classification}

\author[inst1]{Souhail Bakkali\corref{cor1}}
\cortext[cor1]{Corresponding author at: L3i, La Rochelle University, La Rochelle, France}
\ead{souhail.bakkali@univ-lr.fr}
\author[inst1]{Zuheng Ming}
\ead{zuheng.ming@univ-lr.fr}
\author[inst1]{Mickael Coustaty}
\ead{mickael.coustaty@univ-lr.fr}
\author[inst2,inst3]{Marçal Rusiñol}
\ead{marcal@allread.ai,marcal@cvc.uab.cat}
\author[inst3]{Oriol Ramos Terrades}
\ead{oriolrt@cvc.uab.cat}

\affiliation[inst1]{organization={L3i, La Rochelle University},
            city={La Rochelle},
            country={France}}
\affiliation[inst2]{organization={AllRead MLT},
            city={Barcelona},
            country={Spain}}
\affiliation[inst3]{organization={CVC, Universitat Autonoma de Barcelona},
            city={Barcelona},
            country={Spain}}

\begin{abstract}

  \textcolor{black}{Multimodal learning from document data has achieved great success lately as it allows to pre-train semantically meaningful features as a prior into a learnable downstream task. In this paper, we approach the document classification problem by learning cross-modal representations through language and vision cues, considering intra- and inter-modality relationships. Instead of merging features from different modalities into a joint representation space, the proposed method exploits high-level interactions and learns relevant semantic information from effective attention flows within and across modalities. The proposed learning objective is devised between intra- and inter-modality alignment tasks, where the similarity distribution per task is computed by contracting positive sample pairs while simultaneously contrasting negative ones in the joint representation space}. Extensive experiments on public document classification datasets demonstrate the effectiveness and the generality of our model on low-scale and large-scale datasets.
\end{abstract}


\end{frontmatter}

\section{Introduction}
\label{sec:introduction}

\begin{figure*}[t!]
\centering
  \centerline{\includegraphics[width=.90\linewidth]{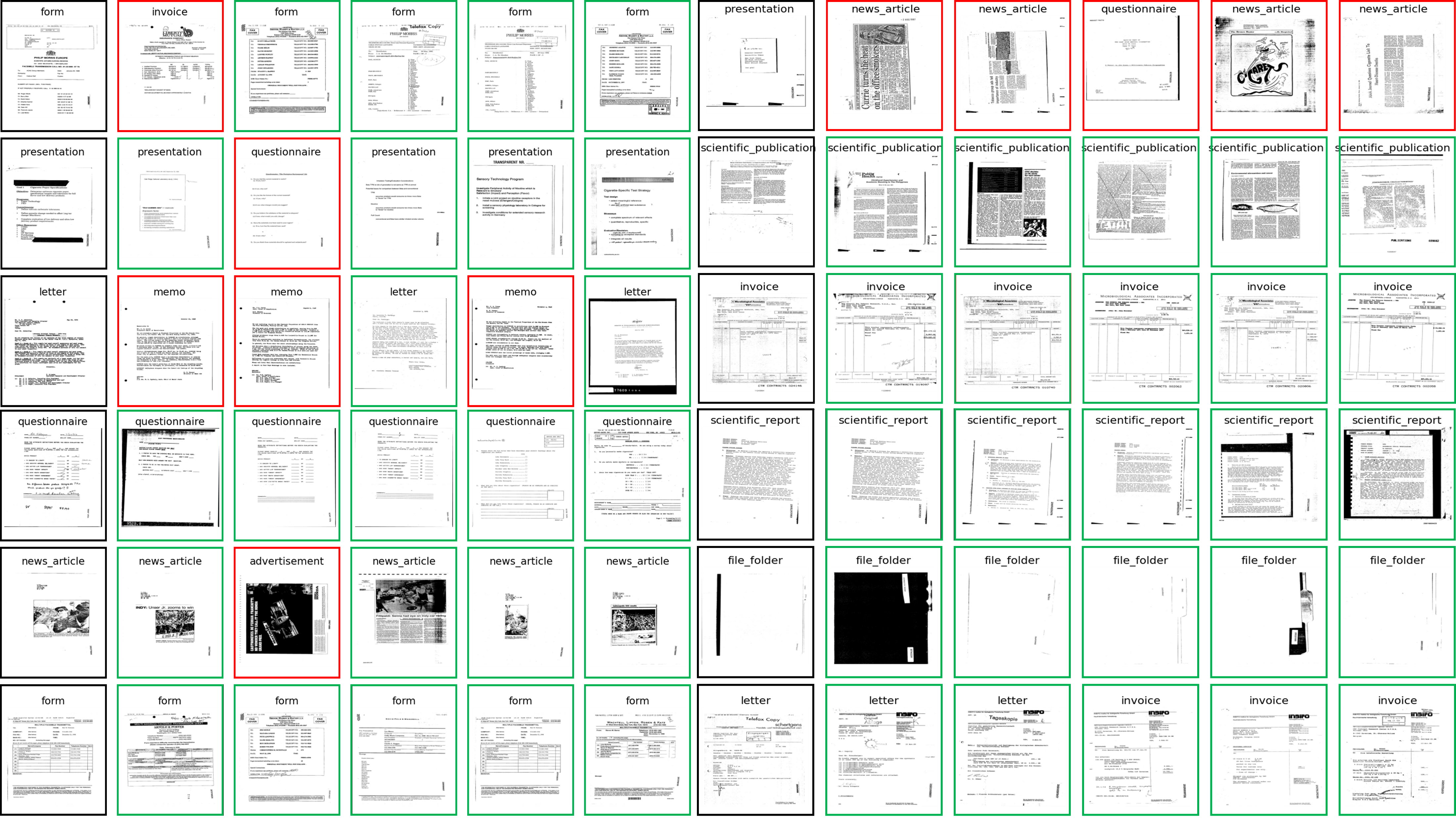}}
    \caption{\textcolor{black}{Document samples from the categories of the RVL-CDIP dataset which show the high intra-class structural variability and low inter-class discrimination between the different categories of document data. Samples from different classes are shown with a red border, while samples from the same class are shown with green border.}}
    \label{fig:dataset}
\end{figure*}

The research field of document image classification brings various challenges for multimodal researchers given the heterogeneity of document data and the contingency often found between its different modalities. Intuitively, documents are natively multimodal. They require multimodal reasoning over multimodal inputs (\eg visual, textual, and layout) which are approximated by combining visual-textual information as two coherent and complementary signals that can be further enhanced with layout information. These documents may be presented in a diverse set of sources such as handwritten text, tables, forms, figures, multi-column layouts, plain text, curved text, and exotic fonts as displayed in the \figurename~\ref{fig:dataset}. Due to the different visual styles, understanding documents visually encounters the problem of low inter-class discrimination, and high intra-class structural variations between the different categories of document data. In general, some documents contain abundant visual information such as reports, and scholarly articles, in which case a stronger emphasis on the semantic meaning of language is more helpful. For instance, some types of documents such as handwriting are mainly not recognizable by OCR algorithms, which lead to losing textual information, and thus, semantic meaning. Then, the visual information within the image regions of the document should be strongly emphasized. Therefore, handling the semantic and stylistic variability in documents is challenging to computational models that are trained mostly on natural images. \textcolor{black}{Furthermore, multimodal reasoning allows to integrate information from language and vision modalities, to reason about the structure of the documents (\eg how the accompanying figures support the text), and to gather the relevant semantic information from the text corpus (\eg how to distinguish between a letter and an email), to finally gather the most important information within the common representation space for decision-making.} 

\textcolor{black}{Most pre-training models rely on training huge datasets to learn a good representation for downstream document applications. However, the shortcomings of the preceding pre-training approaches are three-fold. First, the semantic structure of the document is not only determined by the text within it but also by the visual features (\ie tables, font size and style, figures, etc). For semi-structured documents such as forms and receipts, semantic regions rely more heavily on their surrounding contexts. Given this assumption, page-level pre-training is more preferred than word-level pre-training which allows us to better capture the global information between different modalities guided by the model itself, instead of limiting the model to learn the word-level local features with the given relative position encoding of the words, as well as to learn the multimodal interactions in a more informative and complete way with less possible errors, which occur while using the relative position encoding as visual features (\eg the irregular artistic text in advertisement documents).
Third, during inference, the vision-language sample pairs need to be fed to the fusion modules to calculate the prediction scores in order to perform the document classification task, which remains computationally expensive.} \textcolor{black}{Therefore, the main challenge remains in aligning different modalities without using millions of document samples (320k instead of 11M) as most of pre-trained models. Hence, accessing labels in the pre-training stage makes it possible to have a more adapted and lighter pre-training, which lead to fewer computing resources.}

To address the semantic gap and the lack of closer interactions between image regions and text sequences within and across vision-language modalities, we propose a cross-modal contrastive vision-language pre-training model by learning cross-modal representations as a prior in a unified pre-training network. To encourage cross-modal learning, we model intra- and inter-modality representations between the cues of the vision-language modalities in the pre-training stage. We design an inter-modality cross-attention module denoted as (InterMCA) to capture relevant features from image regions and semantic meaning from text sequences. We aim to ensure that features from vision and language modalities map to closer points in the joint embedding space. Nevertheless, existing cross-modal document understanding approaches lack an explicit measure which ensures that similar features from the same modality stay close in the joint embedding space. We assume that if similar features from the same category of each modality map to distant points in the joint embedding space, then the embeddings generated within vision and language modalities will lack semantically enriched information, and thus, will generalize badly for downstream tasks. As a remedy, we introduce intra-modality representation which is carried within an intra-modality self-attention module denoted as (IntraMSA), which is devoted to constructing intra-modality relations within each modality according to the self-attention weights of image regions and text sequences.

Moreover, leveraging cross-modal relations through InterMCA and IntraMSA attention modules require a cross-modal learning objective. In the pre-training stage, we propose to pre-train the network with a combinatorial cross-modal contrastive learning loss. It aims to simultaneously learn visual-textual features that represent document data in a more efficient manner. For the downstream application, we run uni-modal and multi-modal fine-tuning on top of the pre-trained vision and language encoders to perform document classification. The superior performance on three document datasets demonstrates that the proposed cross-modal learning network, denoted as VLCDoC, can lead to learn meaningful cross-modal representations. The main contributions of this work are summarized as follows:

\begin{itemize}
    \item We design a unified network for cross-modal representation learning. Our network consists of leveraging two flexible extra levels of cross-modal interactions through InterMCA and IntraMSA attention modules, to capture high-level interactions between visual-language cues in document images. The proposed VLCDoC approach shows its superiority over the uni-modal methods.
    \item We propose a cross-modal contrastive learning objective to further explore the relations between vision and language cues. The proposed cross-modal contrastive loss allows to learn and align the feature representations within and across vision-language modalities.
    \item Under a fair comparison setting, our VLCDoC demonstrates a good generality among vision-language based approaches on the benchmark document datasets, and enables to learn robust and domain-agnostic feature representations for document classification. We show that the vision transformer-based architecture used as a backbone of the vision modality in our VLCDoC network can achieve comparable performance when pre-trained on fewer data.
\end{itemize}

\section{Related Work}

\subsection{Multimodal Document Understanding}
\textcolor{black}{\noindent\textbf{Text and Image modalities [T+I]:}
Zhang \etal~\cite{zhang2021trie} proposed a multimodal framework for simultaneous text reading and information extraction for document understanding. Bakkali \etal~\cite{bakkali2020cross, bakkali2020visual} proposed a cross-modal deep network to classify documents in an early fusion manner. Dauphinee \etal~\cite{dauphinee2019modular} constructed a model that uses both the visual information and the textual content of a given document to make a decision in a late fusion manner. 
Also, Bakkali \etal~\cite{bakkali2021eaml} proposed an ensemble self-attention-based mutual learning network that jointly learns text-image features in an end-to-end fashion to classify document images. \\
\noindent\textbf{Text and Layout modalities [T+L]:} LayoutLMv1~\cite{LayoutLMv1} jointly models interactions between text and layout information across document images by adding 2D word position in the language representation to better align the layout information with the semantic representation. 
LILT~\cite{wang2022lilt} fine-tunes on different languages used in pre-training with pre-trained textual models. \\
\noindent\textbf{Text, Layout and Image modalities [T+L+I]:}  LayoutLMv2~\cite{xu2022layoutlmv2} leverages vision, language, and layout modalities in a cross-modal pre-training scheme for a better cross-modality interaction. In LayoutLMv3~\cite{huang2022layoutlmv3}, the authors propose a joint multimodal approach to model the interaction between textual, visual, and layout information in a unified multimodal pre-training network, with different pre-text tasks for a better generality to image-centric and text-centric downstream document AI tasks. Besides, SelfDoc~\cite{li2021selfdoc} exploits cross-modal learning in the pre-training stage to perform a task-agnostic framework to model information across textual, visual, and layout information modalities without requiring document data annotation. In DocFormer~\cite{appalaraju2021docformer}, the authors encourage multimodal interaction using a multimodal transformer architecture to perform visual document understanding.} 

\textcolor{black}{There is a noticeable difference between our proposed method, VLCDoC, and other concurrent works in document image pre-training. The main difference is that our proposed method is page-level, instead of word-level pre-training. Moreover, page-level pre-training allows us to better capture the global information between the vision modality and the language modality by using the entire document. However, the masking-based pre-training strategies that have been widely used in most of the recent pre-training works rely on the layout information of the extracted words along with their relative position encoding within the document image, which leads to learning only the local word-level relationships between the vision modality and the language modality. Nevertheless, in some specific cases of administrative document data like advertisement documents, which are filled by the visual content in most of the regions of the document, even if there are very few text words which can be masked, locating the masked text with artistic style to get an accurate position encoding is also challenging. In this case, capturing the relationships between the vision modality and the language modality by the overall features presented in the whole document are more pertinent than the local information. In this way, we let the model to automatically learn the relationships between the visual features of any region and the corresponding textual features presented in the document image rather than forcing the model to learn and capture the relationships between the given text (masked word) and the corresponding visual features (the location of the masked word within the document image).} 

\subsection{Vision-Language Alignment}
Cross-modal alignment is a broad category of pre-training techniques which aims at mapping text and images into a common space, where semantic similarity across different modalities can be learned by contrastive losses~\cite{yuan2021multimodal, lu2019vilbert, NEURIPS2020_13f320e7}. While dealing with vision-language sample pairs, though individual samples may demonstrate inherent heterogeneity in their content, they are usually coupled with each other based on some higher-level concepts such as their categories. This shared information can be useful in measuring semantics of samples across modalities in a relative manner. Verma \etal~\cite{verma2018cross} analyzed the degree of specificity in the semantic content of a sample in the vision modality with respect to semantically similar samples in the language modality. Krishnan \etal~\cite{krishnan2016matching} measured the similarity score between the words distributions across two document images, by detecting patterns of text re-usages across documents written by different individuals irrespective of the minor variations in word forms, word ordering, layout or paraphrasing of the content. \textcolor{black}{Different from the recent research, the multimodal features are obtained by simply concatenating the text and layout features as the multimodal features but ignore the real cross-modal interactions. However, we propose intra-modality and inter-modality alignment objectives to ensure that samples with semantically similar content stay close in the common space, regardless of the modality. We aim to emphasize the interaction and agreement between visual regions and the semantic meaning of text sequences, as well as to intensify the inner-modality information, by simultaneously preserving the original features and establishing inner-interactions within each modality. Thus, page-level pre-training is necessary to learn the multimodal relationships with the strengths of being able to learn the global information and the overall relationships between the different modalities guided by the model itself, instead of limiting the model to learn the word-level local features with the given relative position embeddings, and also of being able to learn the multimodal interactions in a more informative and complete way with less possible errors which occur while using the coordinates of bounding boxes as visual features. Nevertheless, and to the best of our knowledge, relying on label information of administrative document data is necessary to link the different modalities (\ie the image and the text extracted from the document image) if we would like to conduct the page-level pre-training using the entire document.}

\subsection{\textcolor{black}{Attention Mechanism}}
\textcolor{black}{The attention mechanism was adopted to learn to attend to the most relevant regions of the input space to assign different weights to different regions, and to the most relevant words for the output which often occur at similar positions in the input sequence as introduced by Bahdanau \etal~\cite{Bahdanau2015NeuralMT}. Specifically, self-attention and co-attention learning have been widely applied in multimodal vision-language learning tasks like document understanding, and image captioning~\cite{li2021selfdoc, appalaraju2021docformer, Nam_2017_CVPR}, aiming at learning the internal relations in a text sentence or in an image. To model the internal relationships among different modalities, we adopt the contextualized attention mechanism from NLP ~\cite{vaswani2017attention}
to improve the location accuracy of a document image region in the vision modality for the desired text sequence in the language modality. 
Our proposal highlights cross-modal co-attention (InterMCA), and internal self-attention (IntraMSA) mechanisms which are integrated in VLCDoC.}

\section{Methodology}
\begin{figure*}[t]
\centering
  \centerline{\includegraphics[width=.90\linewidth]{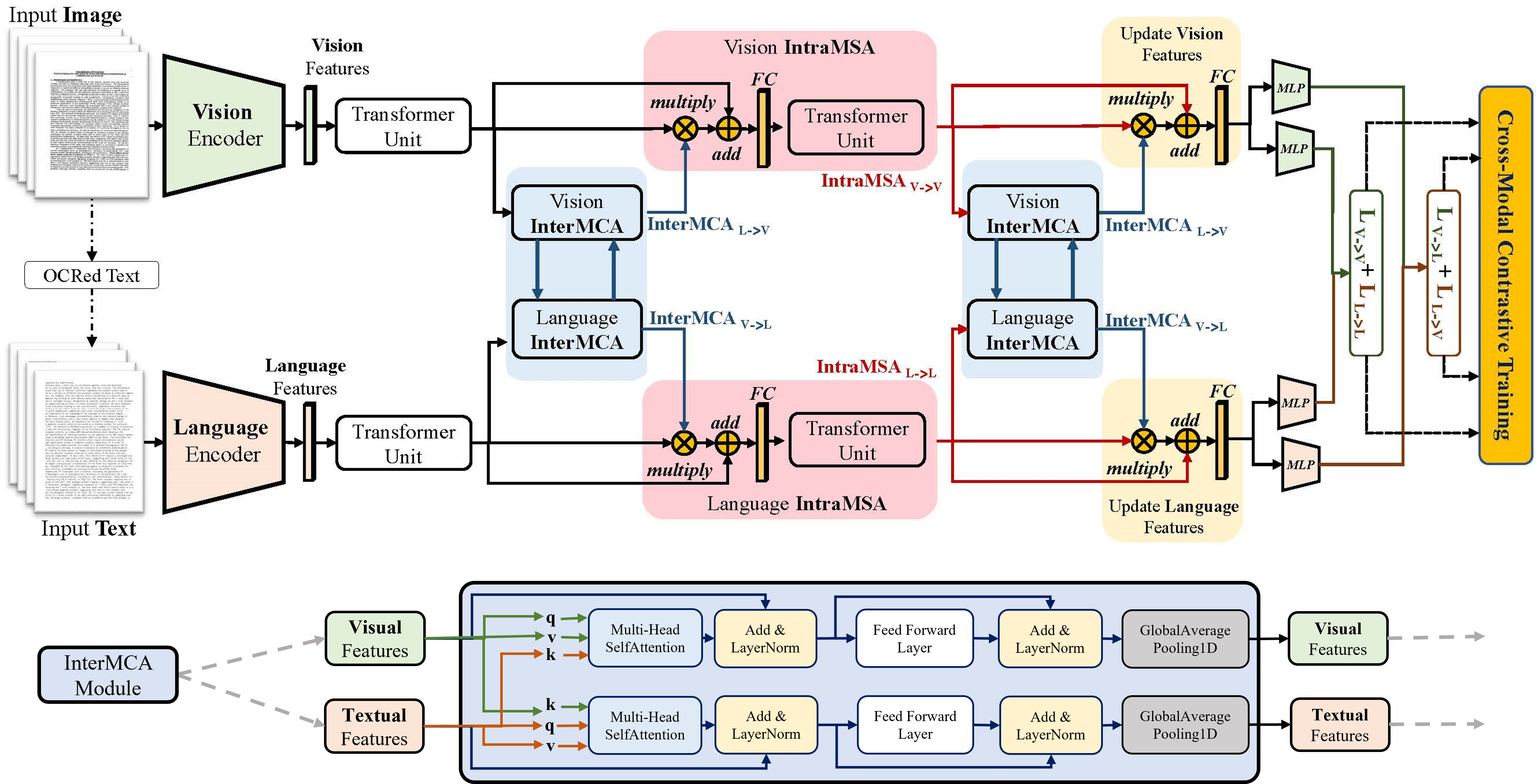}}
    \caption{\textcolor{black}{Overview of the proposed cross-modal contrastive learning method. The network is composed of InterMCA and IntraMSA modules with flexible attention mechanisms to learn cross-modal representations in a cross-modal contrastive learning fashion.}}
    \label{fig:Figure_01}
\end{figure*}

\figurename~\ref{fig:Figure_01} shows the overall architecture of the proposed cross-modal network named VLCDoC. It is an encoder-only transformer-based architecture trained in an end-to-end fashion. It has two main modalities to perform visual-textual feature extraction. VLCDoC enforces deep multimodal interaction in transformer layers using a cross-modal attention module. The VLCDoC architecture network consists of two main schemes: one contrastive learning branch for cross-modal representation learning, and one cross-entropy learning branch for classifier learning. This feature learning strategy aims to learn a feature space which has the property of intra-class compactness and inter-class separability, while the classifier learning branch is expected to learn a domain-agnostic classifier with less bias based on the discriminative features obtained from the encoder branch.
\subsection{Model Architecture}
\subsubsection{Visual Features}
To extract visual embeddings, we follow the original pre-trained vision transformer architecture ViT-B/16~\cite{dosovitskiy2021image} as a backbone. Let $v_{visn} \in \mathbb{R}^{H \times W \times C}$ be the document image. We reshape it into a sequence of flattened $2D$ patches $v_{{visn}_{p}} \in \mathbb{R}^{N \times (P^{2}\cdot C)}$, where $(H, W)$ is the resolution of the document image, $C=3$ is the number of channels, $(P, P)$ is the resolution of each document patch, and $N=HW/P^2$ is the resulting number of patches, which serve as the input sequence length for the transformer encoder. The patches obtained are then flattened and mapped to $d$ dimensions as the hidden embedding size. The resulting visual embeddings are then represented as ${V} = v^{i}_{visn} \in \mathbb{R}^{d_{visn}}$, where $d_{visn}$ is a $2D$ vector.

\subsubsection{Textual Features}
\begin{figure*}[t!]
\centering
  \centerline{\includegraphics[width=.90\linewidth]{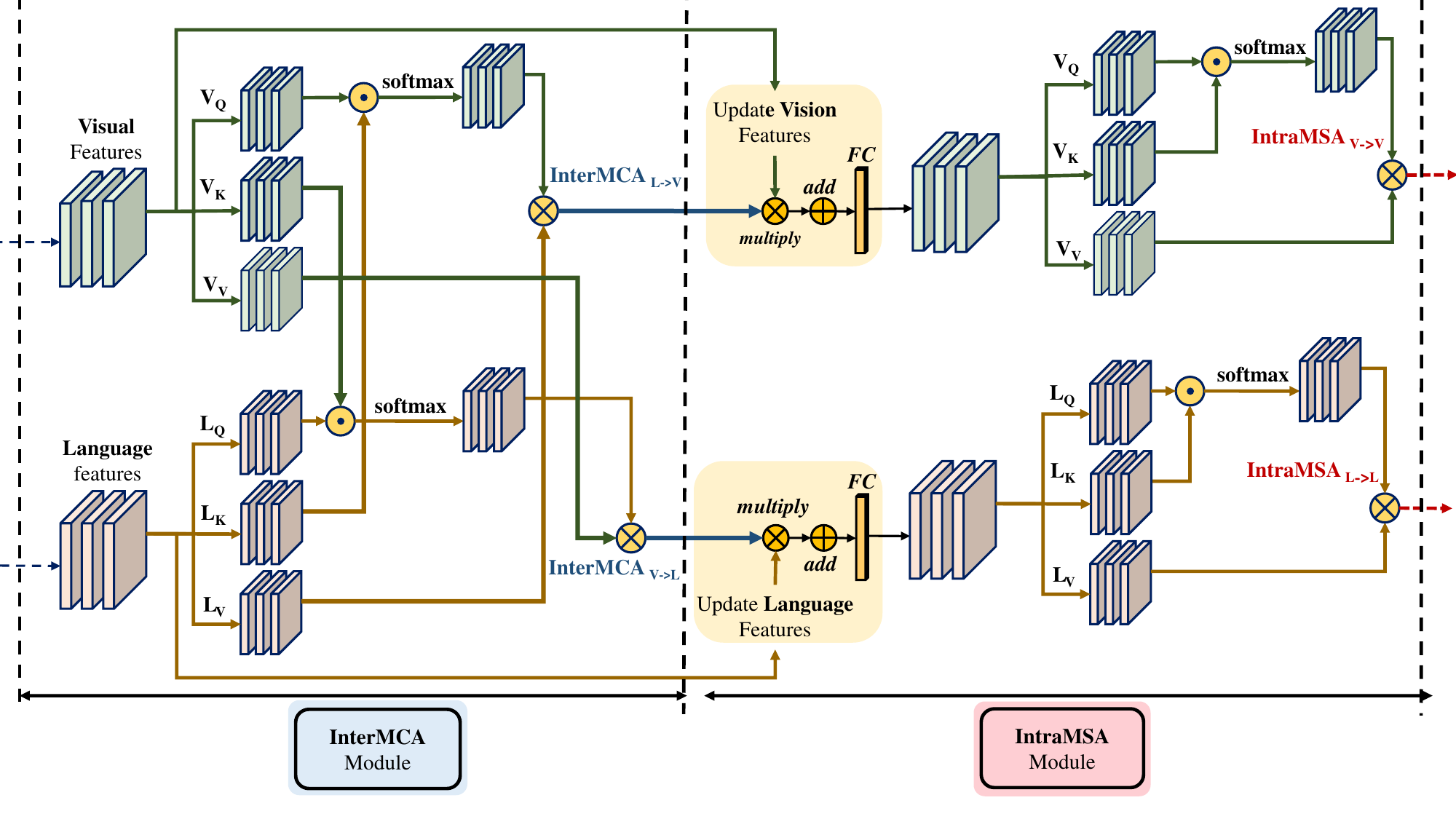}}
    \caption{Illustration of the InterMCA and IntraMSA attention modules. The visual-textual features are transformed into query, key, and value vectors. They are jointly leveraged and are further fused to transfer attention flows between modalities to update the original features.}
    \label{fig:Figure_02}
\end{figure*}
To extract textual embeddings, we first extract the text $t_{lang}$ within document images via an off-the shelf optical character recognition (OCR) system, \eg Tesseract OCR\footnote{\url{https://github.com/tesseract-ocr/tesseract}}. The input sequences extracted with the OCR are further fed into the pre-trained BERT$_{Base}$ uncased encoder~\cite{devlin2019bert}. The resulting textual embeddings are then represented as ${T} = t^{i}_{lang} \in \mathbb{R}^{d_{lang}}$, where $d_{lang}$ is a $2D$ vector of the same size as $d_{visn}$. This way, we ensure that the visual and the textual embeddings are of the same shape.

\subsection{Cross-Modal Alignment}
In this subsection, we introduce the InterMCA and IntraMSA attention modules that capture intrinsic patterns by modeling the inter-modality and intra-modality relationships for image regions and texts. Specifically, our proposed attention modules are transformer-based architectures as in~\cite{vaswani2017attention}. It consists of a multi-head self-attention sub-layer, and a position-wise feed-forward sub-layer $f_{FF}$. Meanwhile, residual connections followed by the layer normalization $f_{LN}$ are also applied around each of the two sub-layers. In the multi-head self-attention sub-layer, the attention is calculated $h$ times, making it to be multi-headed. This is achieved by projecting the queries $\mathcal{Q}$, keys $\mathcal{K}$, and values $\mathcal{V}$ $h$ times by using different learnable linear projections. 

\subsubsection{Inter-Modality Alignment}
The inter-modality cross-attention module InterMCA aims to enhance the cross-modal features by embracing cross-modal interactions across image regions and texts. This module aims to transfer the salient information from one modality to another as illustrated in the \figurename~\ref{fig:Figure_02}. Let $\textbf{V}^l = \{v_{1}, v_{2}, ..., v_{m}\}$, $\textbf{L}^l = \{l_{1}, l_{2}, ..., l_{m}\}$ be the sets of intermediate visual and textual features at the $l$-th layer of the vision and language modalities respectively, where $v_{i} \in \mathbb{R}^{1 \times d_{f}}$, $l_{i} \in \mathbb{R}^{1 \times d_{f}}$, and $\textbf{V} \in \mathbb{R}^{m \times d_{f}}$, $\textbf{L} \in \mathbb{R}^{m \times d_{f}}$. Note that the visual-textual features have the same dimensional feature vector $d_{f}$. To accomplish cross-modal interaction, we apply at first dot-product attention to combine the queries of each modality with the keys of the other. The weighted sum of the value of each modality is computed as:
\small
\begin{align}
    \displaystyle 
        \textbf{InterMCA}_{\textbf{L} \to \textbf{V}}(\textbf{V}^{l}) &= \softmax\left(\frac{\mathcal{Q}_{\textbf{V}^l}\mathcal{K}_{\textbf{L}^l}^\top}{\sqrt{d_{k}}}\right)\mathcal{V}_{\textbf{L}^l}
    \label{eq:01} \\
    \displaystyle 
        \textbf{InterMCA}_{\textbf{V} \to \textbf{L}}(\textbf{L}^{l}) &= \softmax\left(\frac{\mathcal{Q}_{\textbf{L}^l}\mathcal{K}_{\textbf{V}^l}^\top}{\sqrt{d_{k}}}\right)\mathcal{V}_{\textbf{V}^l}
    \label{eq:02}
\end{align}
\normalsize
In this way, we emphasize the agreement between the visual regions and the semantic meaning of texts. The attention weights are then sent into the feed-forward sub-layer. Finally, we get the output features of the next layer of the vision modality $\textbf{V}^{l+1}$ computed as: 
\small
\begin{align}
    \displaystyle
        \textbf{V}_{Att}^l &= f_{LN_{\textbf{V}}}(\textbf{InterMCA}_{\textbf{L} \to \textbf{V}}(\textbf{V}^l) + \textbf{V}^l) \\
    \displaystyle
        \textbf{V}^{l+1} &= f_{LN_{\textbf{V}}}(f_{FF}(\textbf{V}_{Att}^l) + \textbf{V}_{Att}^l)
\end{align}
\normalsize
Similarly, the output features $\textbf{L}^{l+1}$ of the language modality are computed:
\small
\begin{align}
    \displaystyle
        \textbf{L}_{Att}^l &= f_{LN_{\textbf{L}}}(\textbf{InterMCA}_{\textbf{V} \to \textbf{L}}(\textbf{L}^l) + \textbf{L}^l) \\
    \displaystyle
        \textbf{L}^{l+1} &= f_{LN_{\textbf{L}}}(f_{FF}(\textbf{L}_{Att}^l) + \textbf{L}_{Att}^l)
\end{align}
\normalsize
Further, the outputs of each vision and language InterMCA modules are subsequently fed into the vision and language IntraMSA modules.

\subsubsection{Intra-Modality Alignment}
The IntraMSA attention module illustrated in the \figurename~\ref{fig:Figure_02}, aims to update the vision and language information and to capture inner-modality attention weights. For each modality, the information is updated according to a feature fusion scheme. At first, we perform element-wise product to the attention flow $\textbf{V}^{l+1}$ with the the visual region features $\textbf{V}^{l}$, then after a residual connection, features are fused by a linear additive function to yield the final updated visual information. To keep the dimension of the updated information consistent, a fully connected $f_{FC}$ layer is employed. The updated textual information is computed likewise, following the equations:
\small
\begin{align}
    \displaystyle
        \hat{\textbf{V}} &= f_{FC}((\textbf{V}^{l+1} \odot \textbf{V}^{l}) + \textbf{V}^{l}) \\
    \displaystyle
        \hat{\textbf{L}} &= f_{FC}((\textbf{L}^{l+1} \odot \textbf{L}^{l}) + \textbf{L}^{l})
\end{align}
\normalsize
After updating original features based on cross-modal interactions, these features
are fed into the transformer unit to intensify the inner-modality information, to preserve the original features and to establish inner-interactions simultaneously. Following the Equations~\ref{eq:01},~\ref{eq:02}, we have: 
\small
\begin{align}
    \displaystyle
        \textbf{IntraMSA}{_{\textbf{V} \to \textbf{V}}} &= \softmax\left(\frac{\mathcal{Q}_{\hat{\textbf{V}}^l}\mathcal{K}_{\hat{\textbf{V}}^l}^\top}{\sqrt{d_{k}}}\right)\mathcal{V}_{\hat{\textbf{V}}^l} \\
    \displaystyle
        \textbf{IntraMSA}{_{\textbf{L} \to \textbf{L}}} &= \softmax\left(\frac{\mathcal{Q}_{\hat{\textbf{L}}^l}\mathcal{K}_{\hat{\textbf{L}}^l}^\top}{\sqrt{d_{k}}}\right)\mathcal{V}_{\hat{\textbf{L}}^l}
\end{align}
\normalsize
These two modules can be stacked repeatedly, enabling to explore further latent intra- and inter-modality alignments between image regions and texts.

\subsection{Cross-Modal Contrastive Learning}
We design a vision-language contrastive loss to force samples from language and vision that are semantically related to be closer. 
\begin{figure*}[t!]
\centering
\begin{subfigure}{\textwidth}
  {
  \includegraphics[width=.90\linewidth]{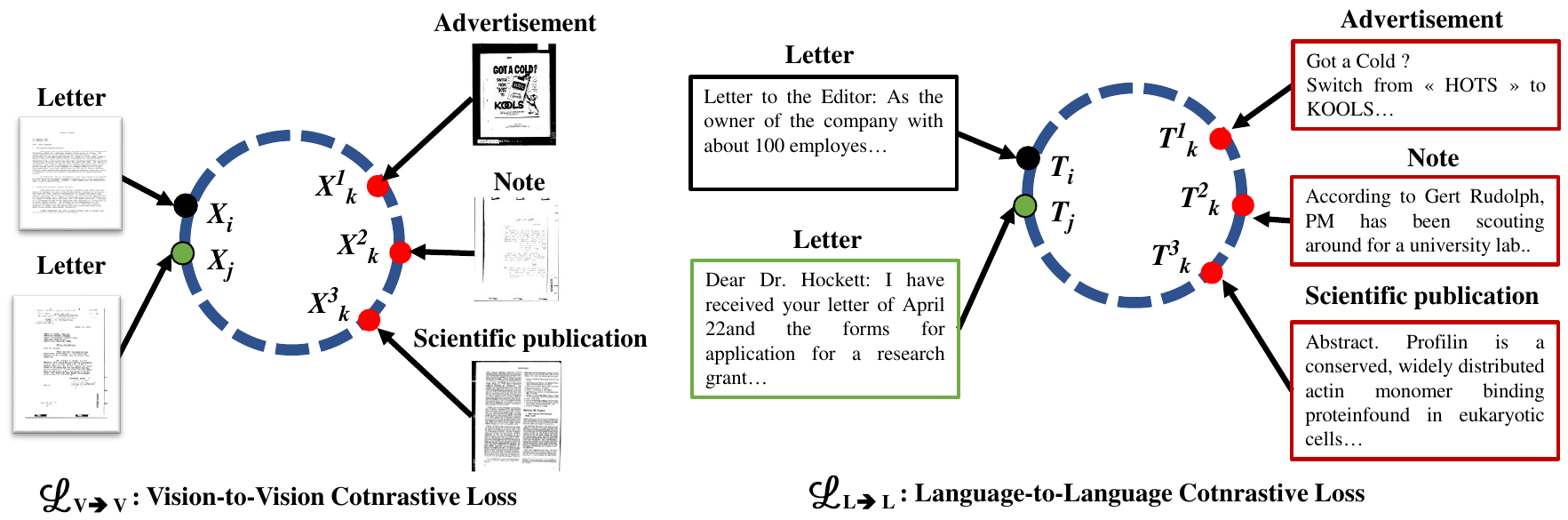}}\quad
  \caption{Intra-modality contrastive learning}
  \label{fig:intra_modal_loss}
\end{subfigure}
\hfill
\begin{subfigure}{\textwidth}
  {
  \includegraphics[width=.90\linewidth]{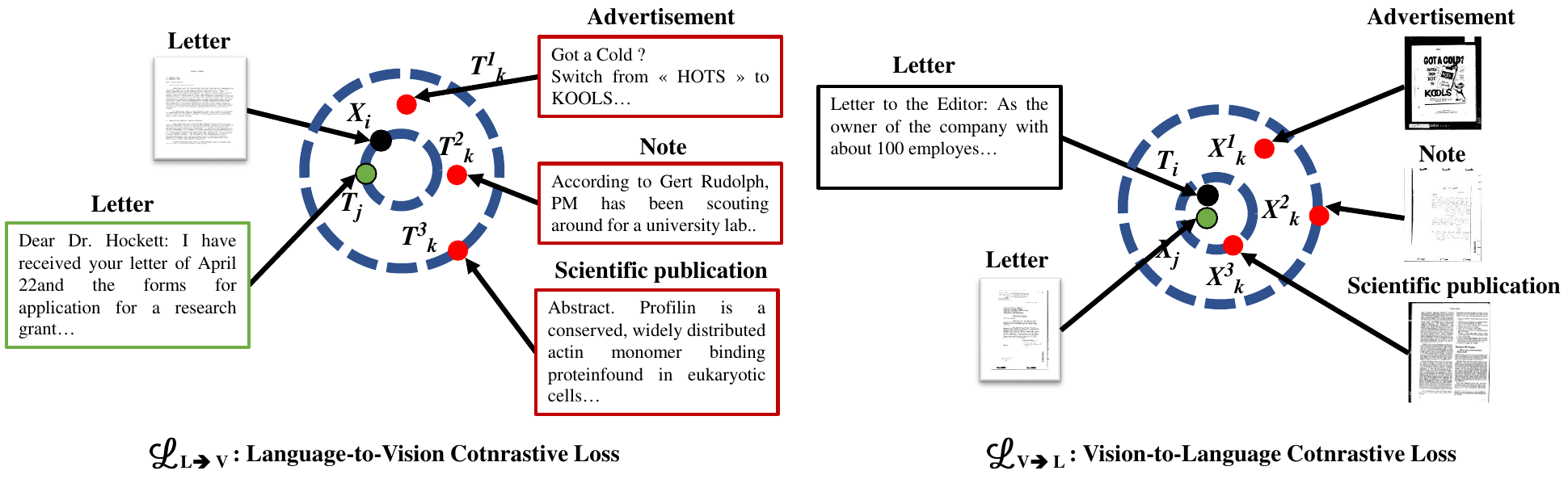}}\quad
  \caption{Inter-modality contrastive learning}
  \label{fig:inter_modal_loss}
\end{subfigure}
\caption{The proposed cross-modal contrastive learning objective}
\label{fig:contrastive_loss}
\end{figure*}
Besides, a projection head is implemented on top of the IntraMSA and InterMCA modules to map the image and text representations into a vector representation so that the two training schemes do not interfere with each other. 
The projection head is implemented as a nonlinear multiple-layer perceptron (MLP) with one hidden layer, as it is more suitable for contrastive learning~\cite{chen2020simple}. 
Then, $L_{2}$ normalization is applied to the visual-textual embeddings so that the inner product between features can be used as distance measurements. In the following parts, we denote cross-modal contrastive learning as CrossCL. 

\subsubsection{Intra-Modality and Inter-Modality Contrastive Learning}
Let $\{\textbf{x}_{i}^+\} = \{x_{j} | y_{j} = y_{i}, i \neq j \}$, $\{t_{i}^+\} = \{t_{j} | y_{j} = y_{i}, i \neq j \}$ be the sets of all positive samples from the same class of an anchor image $x_{i}$ and an anchor text $t_{i}$ respectively, and $\{\textbf{x}_{i}^-\} = \{x_{j} | y_{j} \neq y_{i}, \}$, $\{t_{i}^-\} = \{t_{j} | y_{j} \neq y_{i} \}$ be the sets of the remaining negative samples from other classes within the minibatch N. Not only the pairs ($\textbf{x}_{i}$, $\textbf{x}_{j}$), ($\textbf{t}_{i}$, $\textbf{t}_{j}$) from the same modality should be mapped to a close location in the joint embedding space (intra-modality), but also similar samples $\textbf{x}_{i}$ and $\textbf{t}_{j}$ should be mapped in close proximity (inter-modality). Therefore, the vision modality loss shown on the left of the \figurename s~\ref{fig:intra_modal_loss},~\ref{fig:inter_modal_loss} is computed as: 
\small
\begin{align}
    \displaystyle
        \mathcal{L}_{V} &= \sum_{i=1}^N \mathcal{L}_{V \to V}(\textbf{x}_{i}) + \sum_{i=1}^N \mathcal{L}_{L \to V}(\textbf{x}_{i}) \\
    \displaystyle
        \mathcal{L}_{V \to V}(\textbf{x}_{i}) \!\!&=\!\! \frac{-1}{|\{\textbf{x}_{i}^+\}|}\!\!\sum_{\textbf{x}_{j} \in \{\textbf{x}_{i}^+\}} \!\! \!\!\log \frac{\exp(\textbf{x}_{i} \cdot \textbf{x}_{j} / \tau)}{{\sum_{\textbf{x}_{k}, k\neq i} \exp(\textbf{x}_{i} \cdot \textbf{x}_{k} / \tau)}} \\
    \displaystyle  
        \mathcal{L}_{L \to V}(\textbf{x}_{i}) \!\!&=\!\! \frac{-1}{|\{\textbf{t}_{i}^+\}|} \!\! \sum_{\textbf{t}_{j} \in \{\textbf{t}_{i}^+\}} \!\!\!\! \log \frac{\exp(\textbf{x}_{i} \cdot \textbf{t}_{j} / \tau)}{\sum_{\textbf{t}_{k}, k\neq i} \exp(\textbf{x}_{i} \cdot \textbf{t}_{k} / \tau)}
    \label{eq:13}
\end{align}
\normalsize
where $\cdot$ is the similarity score between example pairs, $\tau$ is a scalar temperature hyper-parameter, N is the minibatch size, $|\{\textbf{x}_{i}^+\}|$ and $|\{\textbf{t}_{i}^+\}|$ denote the number of positive samples of anchors $\textbf{x}_{i}$ and $\textbf{t}_{i}$ respectively. 
Similarly, the language modality loss shown on the right of \figurename s~\ref{fig:intra_modal_loss},~\ref{fig:inter_modal_loss} is computed as: 
\small
\begin{align}
    \displaystyle
        \mathcal{L}_{L} &= \sum_{i=1}^N \mathcal{L}_{L \to L}(\textbf{t}_{i}) + \sum_{i=1}^N \mathcal{L}_{V \to L}(\textbf{t}_{i})
\end{align}
\normalsize
Therefore, the learning objective is based on four contrastive components:
\small
\begin{align}
    \displaystyle
        \mathcal{L}_{CrossCL} &= \mathcal{L}_{V \to V}+ \lambda \mathcal{L}_{L \to V}+\mathcal{L}_{L \to L}+ \lambda \mathcal{L}_{V \to L}
    \label{eq:15}
\end{align}
\normalsize
where $\lambda$ is a hyper-parameter to control inter-modality alignment.

\section{Experiments}
\subsection{Datasets}
\noindent{\textbf{RVL-CDIP.}} \, The RVL-CDIP dataset is a subset of the IIT-CDIP Test Collection presented in~\cite{harley2015evaluation}. It consists of gray-scale labeled documents split into 16 classes. The dataset is split into 320K training documents, 40K documents documents for validation and test sets.\\
\noindent{\textbf{Tobacco-3482.}} \, The Tobacco-3482 dataset is a smaller sample containing 3482 gray-scale document images presented in ~\cite{kumar2014structural}. This dataset is formed by documents belonging to 10 classes not uniformly distributed. For simplicity, we denote the dataset as Tobacco. \\
\noindent{\textbf{NIST Special Database 6.}} \, The Nist-tax form~\cite{dimmick1991nist} dataset is composed of structured forms of 5595 pages of binary, black-and-white images of synthesized documents containing hand-print and split into 20 different tax forms. For simplicity, we denote the dataset as NIST.

\subsection{Experimental Settings}
The proposed VLCDoC method is implemented in Tensorflow with 4 NVIDIA GeForce 12Gb RTX 2080Ti GPU. For the vision modality, documents are resized into a fixed size of (H, W)=(224, 224). The image region feature vector extracted by the ViT-B/16 backbone is of $d_{visn}$=(197, 768). The final vision representation which is fed into the projection head is of dimension $d$=768. As for the textual data, we tokenize the plain text $t_{lang}$ using a word-peace tokenizer to get $t_{tok}$. Each input sequence is expected to start with a $[CLS]$ token, and should end with a $[SEP]$ token. The $t_{tok}$ is then represented as: $t_{tok}= [CLS], t_{{tok}_{1}}, t_{{tok}_{2}}, ..., t_{{tok}_{n}}, [SEP]$, where $n$=197 is the maximum sequence length. For each document, if $n>$197, the input sequence is truncated so that it fits the desired length. For sequences that are shorter than $n<$197, they are padded until they are $n=$197 long. In the pre-training phase, the model is trained using AdamW optimizer with a learning rate of 2e-5, linear warmup ratio to 0.1 and a linear decay. We set the batch size to $64$ and we use the pre-trained weights of both ViT-B/16 and BERT$_{Base}$ uncased backbones. We conduct pre-training for 100 epochs for the RVL-CDIP and Tobacco datasets. We fine-tune our network on 50 epochs for all datasets, we use Adam optimizer with learning rate of 5e-5. 
For Tobacco and NIST datasets, we split the original sets to $80\%$ for training, and $10\%$ for validation and test. The temperature parameter $\tau$ is set to 0.1, and $\lambda$ is set to 0.5. Note that we didn't use any type of data augmentation during pre-training, and we kept the OCRed text as is without any pre- or post-processing. Note that the InterMCA and IntraMSA modules in our method are flexibly stacked two times to enhance the modeling of inter-modality and intra-modality relations during pre-training. We split the query, key, and value vectors of the visual features and textual features into four heads and concatenate the results in different sub-spaces.

\subsection{Ablation Study}
We conduct ablation studies to characterize our VLCDoC network on the low-scale Tobacco dataset. We analyze the following contributions of: i) validating the effectiveness of the proposed InterMCA and IntraMSA attention modules in learning generic cross-modal representations, ii) investigating whether contrastive learning enhances the cross-modal representations, resulting in performance gain in terms of classification accuracy, iii) illustrating the generality and robustness of the proposed VLCDoC network.
\begin{table}[t]
\footnotesize
\centering
\caption{Ablation study on VLCDoC on cross-modality attention components, pre-trained on Tobacco dataset}
{\begin{tabular}{@{}lcccc@{}} 
    \toprule
    Pre-training setting & IntraMSA  &   InterMCA  & \#Params &   Accuracy(\%)\\
    \midrule
     -\textit{vision-only}  &  &  &  & \\
    \midrule
    &        &           &   198M    &   85.71 \\
    &$\surd$ &           &   201M    &   86.66 \\
    &        & $\surd$   &   209M    &   87.20   \\
    &$\surd$ & $\surd$   &    217M   &   \textbf{90.94}\\
    \midrule
     -\textit{language-only}  &  &  &  & \\
    \midrule
    &        &           &   198M    &   86.01 \\
    &$\surd$ &           &   201M    &   86.31 \\
    &        & $\surd$   &   209M    &   87.50   \\
    &$\surd$ & $\surd$   &   217M    &   \textbf{90.62}\\
    \bottomrule
\end{tabular}}
\label{tab:Ablation_Study_Tobacco}
\end{table}

\subsubsection{Effects of Attention Mechanisms}
\begin{figure*}[t]
\centering
\begin{subfigure}{\textwidth}
  {
  \includegraphics[width=.90\linewidth]{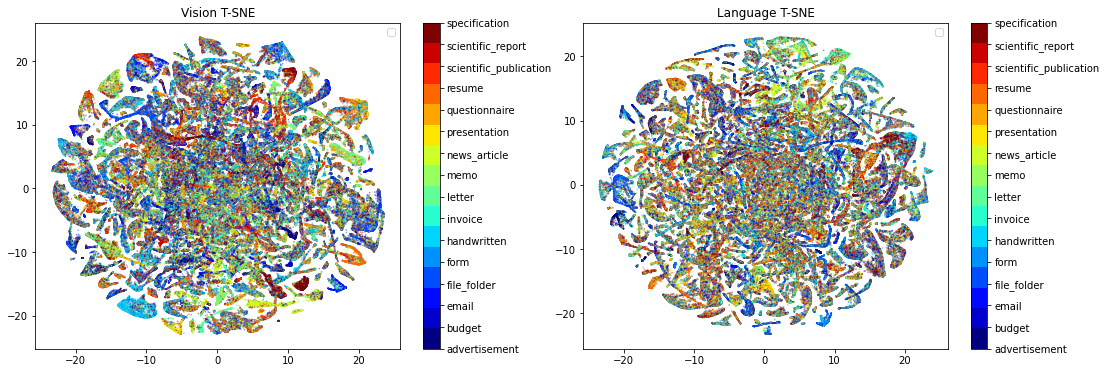}}\quad
  \caption{T-SNE Visualization of VLCDoC without InterMCA and IntraMSA modules}
  \label{fig:T-SNE w/o Attention}
\end{subfigure}
\hfill
\begin{subfigure}{\textwidth}
  {
  \includegraphics[width=.90\linewidth]{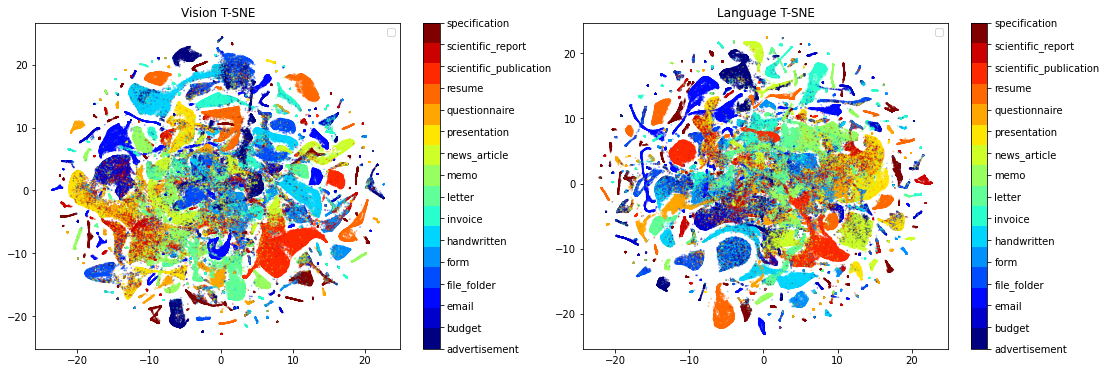}}\quad
  \caption{T-SNE Visualization of VLCDoC with InterMCA and IntraMSA modules.}
  \label{fig:T-SNE w Attention}
\end{subfigure}
\caption{T-SNE vision and language embedding visualization of VLCDoC.}
\label{fig:T-SNE visualization}
\end{figure*}

To investigate the effectiveness of the attention mechanisms used in our VLCDoC model, we evaluate the performance of the learned cross-modal representations w/ and w/o the attention modules. Note that the evaluation protocol is uni-modal based. At first, we consider the scheme where the vision and language modalities are pre-trained independently. In Table~\ref{tab:Ablation_Study_Tobacco}, we observe a significant drop to 85.71\%, and 86.01\% in classification performance when removing both attention mechanisms in the vision and language modalities respectively. When removing only the InterMCA module, we see that our model manages to improve slightly the performance of both modalities to 86.66\% and 86.31\% for the vision-language modalities. Further, removing the IntraMSA and keeping only the InterMCA module enables multimodal pre-training in an end-to-end fashion. The reported results in Table~\ref{tab:Ablation_Study_Tobacco} show that our model gains in performance, and achieves the best performance with 90.94\%, 90.62\% top-1 accuracy for the vision and language modalities.
The improvement of the classification accuracy is attributed to the flexible attention flows adopted in both the InterMCA and IntraMSA modules, which have shown their effectiveness and capability to enhance vision-language relations by capturing the relevant semantic information of images and sentences. \textcolor{black}{\figurename~\ref{fig:T-SNE visualization} illustrates how the document data is arranged in a high-dimensional space without/with IntraMSA and InterMCA attention modules respectively (see \figurename~\ref{fig:T-SNE w/o Attention},~\ref{fig:T-SNE w Attention}), which is based on a T-Distributed Stochastic Neighbor Embedding (T-SNE) algorithm. Therefore, we conclude that both the qualitative and quantitative results demonstrate the effectiveness of cross-modal learning and the importance of both attention modules in learning more effective cross-modal representations during the pre-training stage. }
\begin{table}[t]
\footnotesize
\centering
\caption{Top-1 accuracy (\%) comparison results of our proposed CrossCL loss against the SCL and SSCL losses on the Tobacco dataset}
{\begin{tabular}{@{}lcccc@{}}
\toprule
Model  & Modality & CrossCL (\%) & SCL (\%) \\
\midrule
\textbf{VLCDoC}   &   Vision-only        &   \textbf{90.94}  &   89.88    \\
                  &   Language-only      &   \textbf{90.62}  &   89.29   \\
\bottomrule
\end{tabular}}
\label{tab:ContrastiveLearningComparison}
\end{table}
\subsubsection{Effects of Cross-Contrastive Learning}
The Cross-modal Contrastive Loss (CrossCL) contains two components: intra- and inter-modality alignments. We show the effects of CrossCL on the proposed method against the standard supervised contrastive learning (SCL) loss.
Table~\ref{tab:ContrastiveLearningComparison} shows that the CrossCL loss has a positive impact on the results. The VLCDoC with CrossCL loss yields the best performance gain compared to VLCDoC with the SCL loss. This indicates the importance of CrossCL by enforcing the compactness of intra-class representations (intra-modality), while separating inter-class features by contrasting positive and negative sample pairs within and across each modality. Note that, as described in Equation~\ref{eq:15}, the CrossCL can be vision cue-based or language cue-based, thus we have two different CrossCL presented in Table~\ref{tab:ContrastiveLearningComparison}.
\begin{table}[t]
\footnotesize
\centering
\caption{Cross-dataset test on datasets with different size and document types. Tob, RVL, and Nist denote Tobacco, RVL-CDIP, and Nist-tax form benchmark datasets. Tob $\to$ RVL denotes pre-train on Tobacco, and test on RVL-CDIP.}
{\begin{tabular}{@{}lccccc@{}}
\toprule
\multicolumn{1}{c}{Model}  && \multicolumn{3}{c}{Accuracy (\%)} \\
\midrule
&& Tob $\to$ RVL & RVL $\to$ Tob & RVL $\to$ Nist \\ 
\midrule
\textit{vision-only modality} &  &  \\
 - EAML~\cite{bakkali2021eaml} && 78.89    &   84.82 & - \\
 - \textbf{VLCDoC}              &&  \textbf{79.04} & \textbf{89.73}   &  \textbf{99.99}   \\     
\midrule
\textit{language-only modality} &  &  \\
 - EAML~\cite{bakkali2021eaml} &&  79.06   &  83.72 & -  \\
 - \textbf{VLCDoC}              &&   \textbf{81.96}     &   \textbf{89.88} & \textbf{99.99}  \\
\bottomrule
\end{tabular}}
\label{tab:FineTuningTob}
\end{table}
\subsubsection{Cross-Dataset Test}
To illustrate the generality and the robustness of the learned cross-modal features, we validate our VLCDoC model on document classification datasets with different size and document types. We refer as the cross-dataset test to the process of pre-training our VLCDoC on dataset $A$, and fine-tune it and test it on dataset $B$. The motivation behind is to confirm whether our model displays a good generality in terms of the document classification task. Since there is no publicly available cross-document datasets for this specific task, we evaluate the ability of our model to perform document classification on a new set of documents that had not been seen by our model during the pre-training phase. As denoted in Table~\ref{tab:FineTuningTob}, RVL-CDIP$\to$Tobacco denotes that pre-training is firstly conducted on the RVL-CDIP dataset, then fine-tuning is conducted on the Tobacco dataset. Finally, the inference phase is conducted on the Tobacco dataset as well. Note that during the fine-tuning stage, we only train linear classifiers on top of the final embeddings of the pre-trained vision and language encoders, with the parameters of the rest of the layers freezed. Thus, even though the document categories are different between the dataset $A$ used for pre-training and test dataset $B$ used for fine-tuning and test, we can still evaluate our model on dataset $B$. As such, we compare our model with the related work EAML~\cite{bakkali2021eaml}. We first pre-train the model on Tobacco dataset, then we conduct fine-tuning and test on the RVL-CDIP dataset. The reported results in Table~\ref{tab:FineTuningTob} show that we slightly outperform EAML~\cite{bakkali2021eaml} on both vision and language modalities. Even-though EAML is an ensemble network trained with a different setting, based on vision, language, and fusion modalities, the results confirm that our model benefits from cross-modal pre-training with small amount of document data, achieving better performance with only vision and language modalities. 
Following similar protocol, we pre-train our encoder on RVL-CDIP, and then conduct fine-tuning and test on Tobacco and NIST datasets with fewer document data. We clearly see that our model outperforms the work EAML with a significant margin of 4.91\% and of 6.16\% for vision and language modalities respectively. As for NIST dataset, the results achieve 99.99\% classification accuracy for both modalities. These results demonstrate that our model displays a good generality which enables to learn a robust and domain-agnostic feature representation for classifying documents with different document types and document data size.
\begin{figure*}[t]
\centering
  \centerline{\includegraphics[width=.90\linewidth]{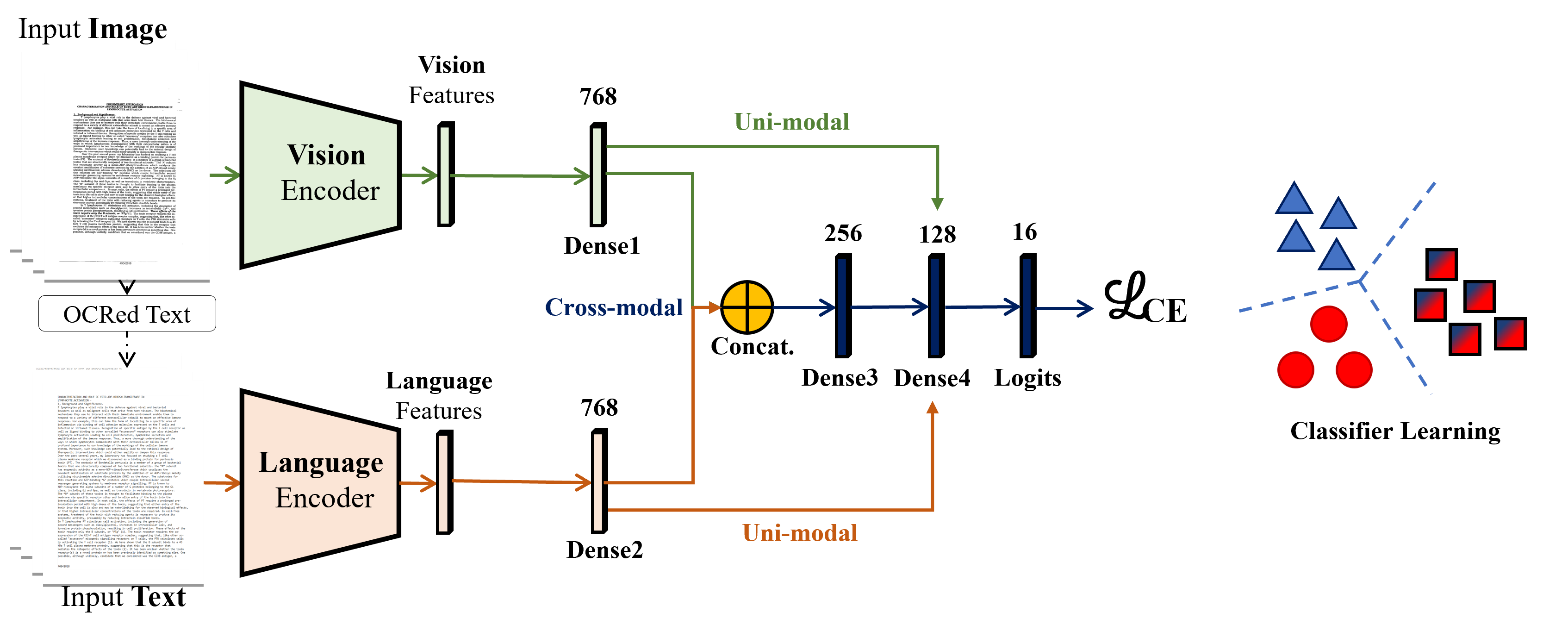}}
    \caption{\textcolor{black}{Overview of the process followed when a visual or textual query is sent to the framework. To perform document classification, we add a set of fully connected layers on top of the pre-trained vision and language encoders with the last layer used as a classifier during the fine-tuning phase, where $L_{CE}$ is the cross-entropy loss. The process of document classification is computed as follows: In the multi-modal fine-tuning phase, we concatenate the layers of each modality (Dense1 for vision data, Dense2 for language data) and pass the multi-modal information through (Dense3, Dense4). Meanwhile, in the uni-modal fine-tuning phase, the connection between layers Dense1 and Dense2 in the concatenation layer is skipped. Instead, the connection is made directly to the Dense4 layer.}}
    \label{fig:Figure_06}
\end{figure*}
\begin{table}[t]
\footnotesize
\centering
\caption{Top-1 accuracy (\%) comparison results of different document classification methods evaluated on the of RVL-CDIP dataset. V+L denotes vision+language modalities}
{\begin{tabular}{@{}lcccc@{}}
\toprule
Method  &  Train Data & Accuracy(\%) & \#Params\\
\midrule
\textit{vision methods} \\
\midrule
VGG-16~\cite{afzal2017cutting}     & 320k & 90.31 & 138M \\
ResNet-50~\cite{afzal2017cutting} & 320k & 91.13 & -\\
Ensemble~\cite{das2018document}        & 320k & 92.21 & -\\
DiT$_{Base}$~\cite{li2022dit} & 42M & 92.11 & 87M \\
\textcolor{black}{\textbf{VLCDoC} (Vision-only)}  & 320k & \textbf{92.64} & 86M \\
\midrule
\textit{(language+layout) methods}\\
\midrule
BERT$_{Base}$~\cite{devlin2019bert}     & - & 89.81 & 110M \\
RoBERTa$_{Base}$~\cite{liu2019roberta}  & - & 90.06 & 125M \\
\textcolor{black}{\textbf{VLCDoC} (Language-only)} & 320k & \textbf{91.37}   & 110M \\
LayoutLM$_{Base}$~\cite{LayoutLMv1}     & 11M & 91.78 & 113M \\
\midrule
\textit{(vision+language) methods} \\
\midrule
Multimodal~\cite{audebert2019multimodal} & 320k & 90.6 & -\\
Ensemble~\cite{dauphinee2019modular} & 320k & 93.07 & -\\
\textcolor{black}{\textbf{VLCDoC (V+L)}} & 320k & \textcolor{black}{\textbf{93.19}}   & 217M \\
EAML~\cite{bakkali2021eaml} & 320k & 94.44 & -\\
\midrule
\textit{(vision+language+layout) methods} \\
\midrule
SelfDoc ~\cite{li2021selfdoc} & 320k & 92.81 &  -\\
LayoutLM$_{Base}$~\cite{LayoutLMv1} &  11M & 94.42 & 160M \\
TILT$_{Base}$~\cite{powalski2021going} & 1M & 95.25 & 230M \\
LayoutLMv2$_{Base}$~\cite{xu2022layoutlmv2} & 11M & 95.25 & 200M \\
LayoutLMv3$_{Base}$~\cite{huang2022layoutlmv3} &  11M & 95.44 & 133M \\
DocFormer$_{Base}$~\cite{appalaraju2021docformer} & 5M & 96.17 & 183M \\
\bottomrule
\end{tabular}}
\label{tab:SOTAComparison}
\end{table}

\subsection{Results}
The comparison between the proposed VLCDoC network and existing methods on the large-scale RVL-CDIP document classification dataset is presented in Table~\ref{tab:SOTAComparison}. The compared methods cover various training strategies with different modalities used to perform document classification. These methods include vision-only, language-only, vision+language, and 
vision+lang-uage+layout methods. \textcolor{black}{Although our VLCDoC network learns feature space with vision and language cues, it uses both uni-modal (either vision or language) and multi-modal (vision+language) modalities to classify document images during fine-tuning and inference phases as illustrated in \figurename~\ref{fig:Figure_06}. Note that for the uni-modal task, it is performed on top of the pre-trained vision and language encoders, by adding a fully-connected layer, and the softmax function to perform the classification task. In Table~\ref{tab:SOTAComparison}, we can see that our VLCDoC model achieves good performance with 92.64\% top-1 accuracy regarding the vision-only modality setting. As for the uni-modal language-only setting, we also achieve good performance of 91.37\% accuracy compared to the language+layout methods with large amount or pre-training data (LayoutLM with 11M vs VLCDoC with 320k). Therefore, in the multi-modal fine-tuning setting, where both vision+language modalities are used, the results reported demonstrate that our proposed VLCDoC (Vision+Language) outperforms both uni-modal modalities with an accuracy of 93.19\% compared to 92.64\% and 91.37\% for vision and language modalities respectively. Meanwhile, it achieves competitive results against the methods that include layout information in the pre-training setting (\eg SelfDoc~\cite{li2021selfdoc}. Finally, the results indicate that a vision encoder-only transformer-based architecture can help achieve compelling results in the uni-modal setting, but still struggles to boost the performance in the multi-modal fine-tuning setting against EAML~\cite{bakkali2021eaml} which is based on a DCNN ResNetInceptionV2 architecture~\cite{szegedy2017inception}.} \textcolor{black}{Therefore, even-though vision transformers have achieved competitive results with DCNNs in visual recognition domain, inductive bias makes them more data-hungry than common DCNNs to avoid over-fitting in downstream applications with less data such as the document understanding domain\cite{wu2021cvt}. Thus, vision, language, and layout-based transformer methods have shown to be effective in learning accurate representations with a large amount of pre-training data of 11M documents, demonstrating the effectiveness of transformers against DCNNs. Nevertheless, our proposed VLCDoC method achieves compelling classification results and reduces the gap between vision, language, and layout-based transformer methods with a low amount of pre-training data, and demonstrates a good generalization ability on unseen document data.}

\section{Conclusion and Future Work}
In this paper, we proposed a novel cross-modal representation learning model for document classification, which models the intra- and inter-modality relations between vision-language cues. We have introduced InterMCA and IntraMSA attention mechanisms which incorporate visual-textual features to further improve the cross-modal representations. We have performed a detailed analysis and evaluation on each module, demonstrating the suitability of the proposed approach. We have demonstrated a good generality of our multimodal transformer-based model to the document classification task, enabling to classify documents in different domains. We will push forward two research lines for the future. On the one hand, we will carry on further research on the integration of a third layout modality in our transformer-based multimodal model. We would like to propose a better solution for layout integration in our vision-language model. On the other hand, we would like to explore new pre-text task strategies to improve document understanding in a pretrain-then-finetune paradigm. Thus, we will further tune our model on different downstream applications related to document AI with more challenging heterogeneous data.

\section{Acknowledgements}
This work has been co-funded by the French National Research Agency (ANR), and partially supported by the Spanish projects PID2021-126808OB-I00, the Catalan project 2021-SGR-01559  and the CERCA Program / Generalitat de Catalunya.

\bibliographystyle{elsarticle-num}
\bibliography{cas-refs} 
\end{document}